\documentclass[letterpaper]{article} 
\usepackage{aaai24}  
\usepackage{times}  
\usepackage{helvet}  
\usepackage{courier}  
\usepackage[hyphens]{url}  
\usepackage{graphicx} 
\usepackage{natbib}  
\usepackage{caption} 
\usepackage{algorithm}
\usepackage[noend]{algpseudocode}

\usepackage{xcolor}
\usepackage{graphicx}
\usepackage{amsmath}
\usepackage{amssymb}
\usepackage{cuted}

\usepackage{booktabs}
\usepackage{mathtools}
\usepackage{makecell}
\usepackage{bm,multirow}
\usepackage{multicol}
\usepackage{subcaption}

\usepackage{arydshln}

\usepackage{newfloat}
\usepackage{listings}
\DeclareCaptionStyle{ruled}{labelfont=normalfont,labelsep=colon,strut=off} 
\floatstyle{ruled}
\newfloat{listing}{tb}{lst}{}
\floatname{listing}{Listing}
\title{Patch-Wise Graph Contrastive Learning for Image Translation}
\author{
	Chanyong Jung\textsuperscript{\rm 1}, Gihyun Kwon\textsuperscript{\rm 1}, Jong Chul Ye\textsuperscript{\rm 1, \rm 2}
}
\affiliations{
    \textsuperscript{\rm 1} Department of Brain and Bio Engineering, KAIST, Daejeon, Republic of Korea \\
    \textsuperscript{\rm 2} Kim Jaechul Graduate School of AI, KAIST, Daejeon, Republic of Korea \\
    
    \{ jcy132, cyclomon, jong.ye \}@kaist.ac.kr
}

\usepackage{bibentry}

\begin{document}

\maketitle

\begin{abstract}
Recently, patch-wise contrastive learning is drawing attention for the image translation by exploring the semantic correspondence between the input  and output images.  
To further explore
the patch-wise topology for high-level semantic understanding, here we exploit the graph neural network to capture the topology-aware features.
Specifically, we construct the graph based on the patch-wise similarity from a pretrained encoder, whose  adjacency matrix is shared to enhance the consistency of patch-wise relation between the input and the output. Then, we obtain the node feature from the graph neural network, and enhance the correspondence between the nodes by increasing mutual information using the contrastive loss. In order  to capture the hierarchical semantic structure, we further propose
the graph pooling. 
Experimental results demonstrate the state-of-art results for the image translation thanks to the semantic encoding by the constructed graphs.

\end{abstract}

\section{Introduction}
\label{sec:intro}

Image-to-image translation task is a conditional image generation task in which the model converts the input image into target domain while preserving the content structure of the given input image. The seminar works of image translation models used paired training setting~\cite{pix2pix}, or cycle-consistency training~\cite{cyclegan} for content preservation. However, the models have disadvantages in that they require paired dataset or need complex training procedure with additional networks. To overcome the problems, later works introduced one-sided image translation by removing the cycle-consistency~\cite{gcgan,distancegan}.

\begin{figure}[!t]
	\includegraphics[width=0.99\linewidth]{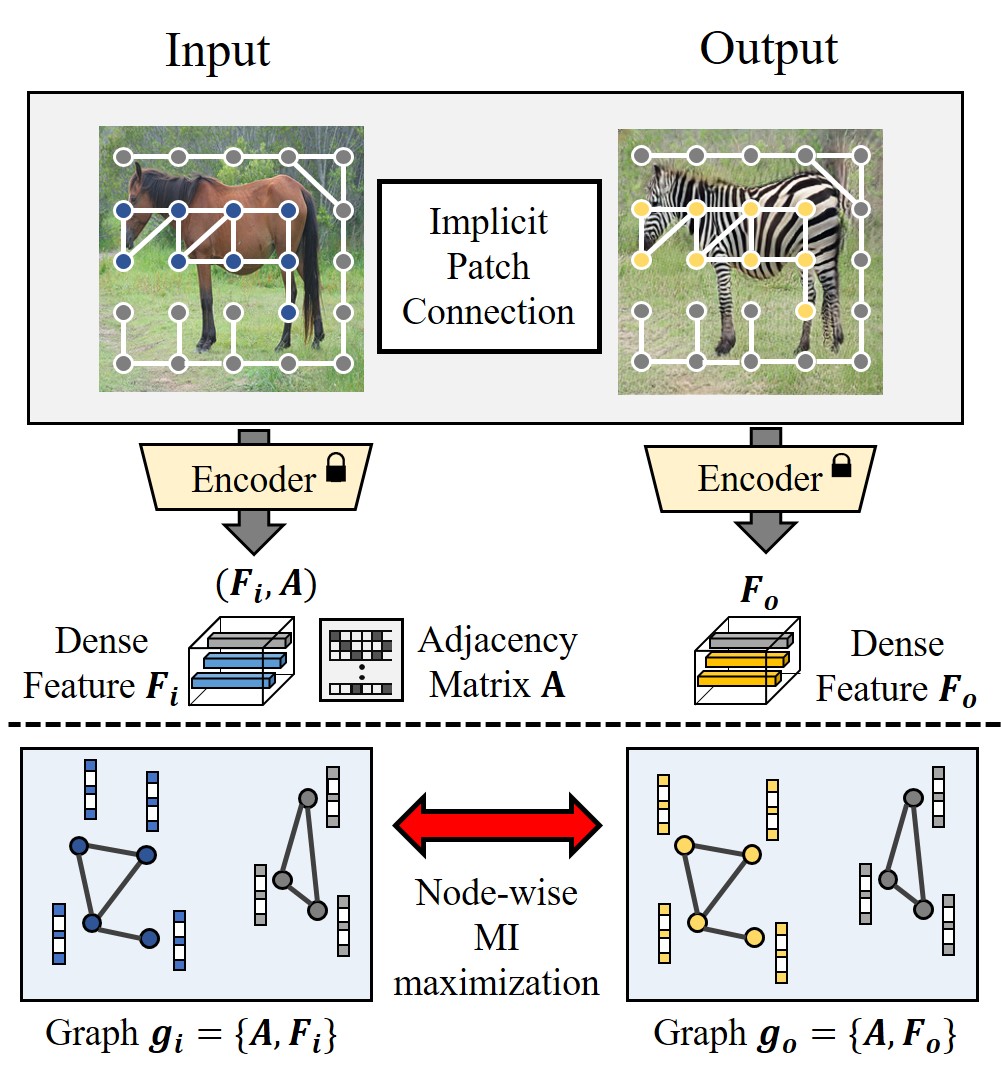}
	\caption{
		The semantic connectivity of input is extracted by the encoder, and shared to construct the graph network. We maximize the mutual information between the nodes. }
	\label{fig:concept}
\end{figure}

Recently, inspired by the success of contrastive learning strategies, Contrastive Unpaired Translation (CUT)~\cite{cut} is proposed to enhance the correspondence between the input and the output images by the patch-wise contrastive learning. 
The patch-wise contrastive learning is further improved by exploring patch-wise relation such as adversarial hard negative samples~\cite{negcut}, patch-wise similarity map~\cite{sesim}, or consistency regularization combined with hard negative mining by patch-wise relation~\cite{HnegSRC}. Although these methods show meaningful improvement in the performance, they still have a limitation in that the previous works focused only on the individual point-wise matching for each pair, which does not consider the topology with the neighbors~\cite{hkd}.

To further explore the semantic relationship between the patches, this paper considers   image translation tasks as topology-aware representation learning as shown in Fig.~\ref{fig:concept}.  Specifically, we propose a novel framework based on the patch-wise graph constrastive learning using the Graph Neural Network (GNN) which is commonly used to extract the feature considering the topological structure. 

Several existing works have utilized GNN to capture topology-aware features for various tasks. Hierarchical representation with graph partitioning is proposed for the unsupervised segmentation~\cite{deepSpectral,tokenCut}, and topology-aware representations~\cite{visionGNN} are extracted based on semantic connectivity between image regions. For knowledge distillation, claimed the {\em holistic knowledge}~\cite{hkd} between the data points is claimed, verifying its effectiveness to encode the topological knowledge of the teacher model.

Despite the great performance in various vision tasks,
none of researches have explored the topology-aware features considering the implicit patch-wise semantic connection for the image-to-image translation tasks. 
Accordingly, here we employ GNN to utilize the patch-wise connection of input image as a prior knowledge for patch-wise contrastive learning. 
Specifically, we  use a pre-trained network to extract the patch-wise features for the input and the output images. Then, we obtain the adjacency matrix calculated by the semantic relation between the patches of the input image, and share it for output image graph. We construct two graphs for the input and the output by the adjacency matrix and the patch features, and obtain the node features by the graph convolution. By maximize the mutual information (MI) between the nodes of input graph and output graph through the contrastive loss, we can enhance the correspondence of patches for the image translation task. 
Furthermore, to extract the semantic correspondence in a hierarchical manner,
we propose to use the graph pooling technique that resembles the attention mechanism. 

Our contributions can be summarized as follows:
\begin{itemize}
	\item We propose a GNN-based framework to capture topology-aware semantic representation by exploiting the patch-wise consistency  between the input and translated output images.
	\item We suggest a method to share the adjacency matrix in order to utilize the patch-wise connection of input image as a prior knowledge for the contrastive learning.  
	\item To further exploit the hierarchical semantic relationship, we propose to use the graph pooling which provides a focused view for the graph.
	\item Experimental results in five different datasets demonstrates the state-of-the-art performance
	by producing semantically meaningful graphs. 
\end{itemize}

\section{Related Works}
\label{sec:related_works}
\paragraph{Patch-Wise Contrastive Learning for Images}
In patch-level view, the image has diverse local semantics. The relational knowledge between the patches embodies the correlation between each region, and is utilized for various image generation tasks. 

For example, patch-wise contrastive relation~\cite{cut, negcut} is utilized for the image translation. Similarly, patch similarity map obtained from pretrained encoder~\cite{sesim} is suggested. Recently, patch-level self-correlation map~\cite{mcl}, query selection module based on patch-wise similarity~\cite{qsAttn}, optimal transport plan by patch-wise cost matrix~\cite{moNCE} are suggested. Also, semantic relation consistency~\cite{HnegSRC} is proposed for the image translation tasks.
Especially, for style transfer, patch-level relation extracted by vision transformer is recently proposed~\cite{splicing,text2live}. The methods utilized the relation between image tokens to preserve the regional correspondence. 
Recently, the consistency of the patch-wise semantic relation between the input and the output images was exploited to further improve the correspondence between the input and the output image~\cite{HnegSRC}.
For style transfer, the consistency of patch-level relation extracted by vision transformer was also studied~\cite{splicing,text2live}.

\begin{figure*}[!t]
	\centering
	\includegraphics[width=0.99\linewidth]{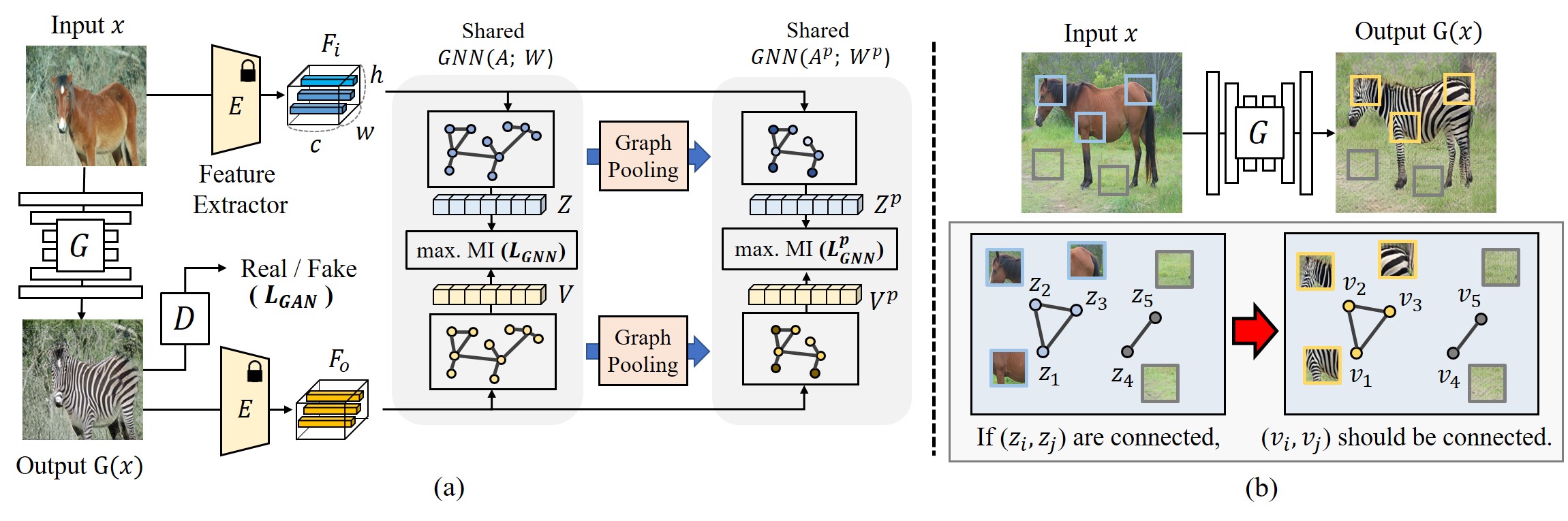}
	\caption{(a) Overall framework of the proposed method. We impose patch-wise regularization by the GNN constructed by the encoder $E$. We extract the node feature $Z, V$ and maximize $I(Z; V)$. Pooled graphs are utilized to focus on task-relevant nodes. (b) The motivation of the proposed approach to use patch-wise connection of input image as the prior knowledge. 
	}
	\label{fig:method}
\end{figure*}

\paragraph{Graph Neural Network}

Graph neural network(GNN) learns the representation considering the connectivity of a graph-structured data~\cite{gcn, tagcn}. Each node feature models the individual data and its relation to the other data points, aggregating the information from the neighbor nodes. 

Thanks to the successes of the GNN to capture the topology-aware features~\cite{lagraph, sep, structPool}, the GNN is actively used in various computer vision tasks.
For example, the GNN is utilized to capture the local features to find image correspondence~\cite{superGlue}, and multi-modal feature for action segmentation in videos~\cite{semantic2graph}. Especially, knowledge distillation method through GNN~\cite{hkd, gkd} is proposed, which is claimed better than conventional contrastive loss, by transferring an additional knowledge on the instance-wise relations.

Recently, the graph constructed by the patch-wise relation was suggested to capture the visual features. The graph partitioning methods are employed for the unsupervised segmentation~\cite{deepSpectral,tokenCut}, where the graph is obtained by the token-wise similarity from the vision transformer. Vision GNN~\cite{visionGNN} is introduced, which have GCN-based architecture to extract the topology-aware representation, and showed its superior performance to the widely used models such as the CNN and the vision transformers.


\section{Method}

Inspired by the previous works, we are interested in exploiting patch-wise relation that represents semantic topology of the image. 
In particular, we focus on the topology-aware features using graph formed by the semantic relation of patches, and explore how the features improve the task performance. 

Specifically, our method is motivated by the consistency of the patch-wise semantic connection of the input and the output images, as shown in Fig.~\ref{fig:method}(b). If the patch features ($z_i, z_j$) have semantic connection in the input image, then the patches ($v_i, v_j$) for the corresponding location of the output should also have the connection. From the motivation, we present a method that  utilizes the topology of patch-wise connection of the input image as a prior knowledge. 

More specifically, we capture the topology-aware patch features by a GNN, where the patch-wise connection is given by the shared adjacency matrix $A$. We then obtain the node features $Z=\{z_i\}_{i=1}^N$ and $V=\{v_i\}_{i=1}^N$ and maximize node-wise MI by the contrastive loss. We also utilize the graph pooling, to maximize the MI within the task-relevant focused view of the graph.  More details follows.

\paragraph{Graph Representation for Image Translation}\label{sec:graph}

We first construct the graph for input image $g_{i} = \left\{ A, F_i\right\}$, where $A$ is adjacency matrix and $F_i$'s are node features that represent the image patches. Specifically, we randomly sample $N$ patch features $f_n \in \mathbb{R}^{c}$ from the dense feature $F=E(x) \in \mathbb{R}^{c \times h \times w }$ which is obtained from the intermediate layer of model $E$, where $c,h,w$ denote the number of color channel, height, and width, respectively.
We set the $N$ features as the nodes for the graph $g_i$ (i.e. $F_i=\left[f_1, …, f_N\right]$).

Then, we obtain the adjacency matrix $A \in \mathbb{R}^{N \times N}$ according to the cosine similarity of the patch features. We connect the patches if the similarity is above the predefined threshold $t$, and disconnect them in otherwise. Specifically, the connectivity $A_{ij}$ for features $f_i, f_j$ is computed by
\begin{align}
	A_{ij} \coloneqq
	\begin{cases}
		1 & \text{ if } \mathrm{cos}(f_i, f_j) \geq t \\ 
		0 & \text{ if } \mathrm{cos}(f_i, f_j) < t
	\end{cases}
\end{align}

\begin{figure}[!t]
	\includegraphics[width=0.99\linewidth]{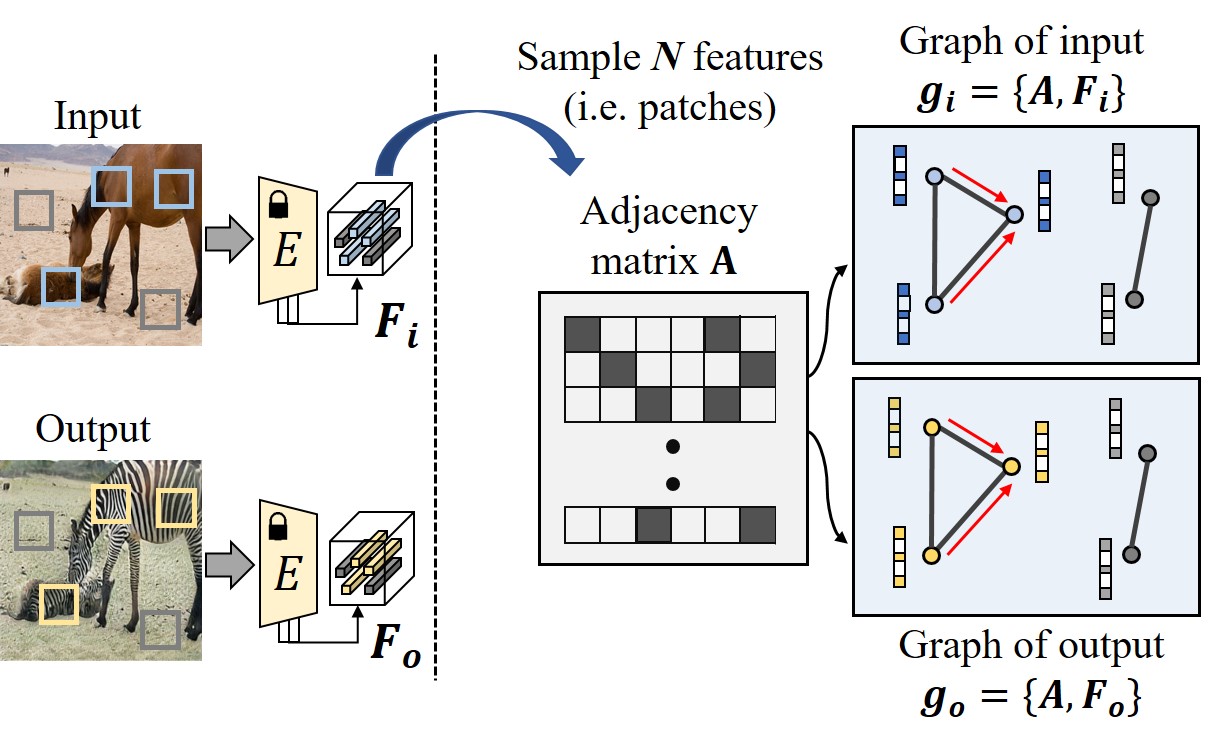}
	\caption{The construction of graphs $g_o, g_i$ with shared adjacency matrix $A$. Each graph extracts $l$-hop features $Z, V$ from the given node $F_i, F_o$.}
	\label{fig:graph}
\end{figure}

We construct the output graph $g_{o} = \{A, F_o\}$ in similar way. We sample $N$ features $f'_n \in \mathbb{R}^{c}$ from the corresponding location of the dense feature $F’=E \circ G(x) \in \mathbb{R}^{c\times h \times \ w}$, and set as the nodes for the graph $g_o$ (i.e. $F_o=\left[f'_1, …, f'_N\right]$ ). To retain the topological
correspondency between the patches, the output graph inherits the adjacency matrix $A$ from the input graph as shown in Fig.~\ref{fig:graph}.

Next, we obtain the graph representation $Z, V$ using Topology Adaptive Graph Convolution Network ~\cite{tagcn} by the graph $g_{o}, g_i$ as follows:
\begin{align}
	Z = \sum_{l=0}^L (\bar{A})^l F_i W_{l} \\
	V = \sum_{l=0}^L (\bar{A})^l F_o W_{l}
\end{align}
where $\bar{A}$ is the normalized adjacency matrix, and $W_{l}$ is the shared parameter for the $l$-th hop. 
We obtain 2-hop representation from the graph (i.e. $L=2$).

Finally, to enforce the topological correspondence between input $X$ and output $G(X)$ for a given generator $G$,
we maximize the mutual information between the nodes $Z, V$ by the infoNCE loss ~\cite{cpc} as follows:
\begin{align}\label{eq:infoNCE}
	L_{GNN}(X, G(X)) = -\frac{1}{N} \sum_{i=1}^N \left[ \log \frac{ \exp (z_i^\top v_i)}{\sum_{j=1}^N \exp (z_i^\top v_j)} \right]
\end{align}
where $z_i, v_i$ are the $i$-th node features from $Z$ and $V$ from $X$ and $G(X)$, respectively.

When $L=0$, the proposed method shrinks to the conventional patch-wise contrastive learning with the projector network $W_0$. In this perspective, our method utilizes the higher-ordered features by the graph aggregation~(i.e. $L>0$), which generalizes the conventional contrastive learning.

\paragraph{Graph Pooling for Focused Attention}
\label{sec:pooling}

We pool the graph nodes to utilize task-relevant focused attention of the graph. In other words, we downsample the nodes by its relevancy to the task, and construct the graph with fewer nodes to focus on the task-relevant nodes.

Specifically, following the top-$K$ pooling ~\cite{graphUnet}, we select $K$ nodes from the $N$ nodes $Z=\left[z_1, …, z_N\right]$ by the similarity score $s_i=p^\top z_i$, where $p$ is the learnable pooling vector. Accordingly, the adjacency matrix $A_p \in \mathbb{R}^{K \times K}$ for the pooled graph is constructed, by excluding the connections with non-selected nodes from the original matrix $A$.
Then, the nodes are weighted by the score followed by sigmoid funcion $\sigma$ as:
\begin{align}
	Z_{p, in} &= \sigma(S)Z \\
	V_{p, in} &= \sigma(S)V
\end{align}
which becomes the input nodes for the pooled graphs. Then, the $L$-hop features are obtained as:
\begin{align}
	Z_{p} &= \sum_{l=0}^L (\bar{A_p})^l Z_{p, in} W_{p,l} \\
	V_{p} &= \sum_{l=0}^L (\bar{A_p})^l V_{p, in} W_{p,l} 
\end{align}
where $W_{p,l}$ is the parameter of the pooled GNN. 
By constructing the pooled graphs $g_i^p=\{A_p, Z_p\}$, $g_o^p=\{A_p, V_p \}$ and obtaining the $l$-hop node feature,
we also employ the infoNCE loss to maximize the MI between the nodes in the pooled graph as follows: 

\begin{align}
	L_{GNN}^p(X, G(X)) = -\frac{1}{K} \sum_{i=1}^K 
	\left[ \log \frac{ \exp(z_{p, i}\top v_{p,i})}
	{\sum_{j=1}^N \exp (z_{p,i}\top v_{p,j})} 
	\right]
\end{align}

\begin{figure}[!t]
	\centering
	\includegraphics[width=0.99\linewidth]{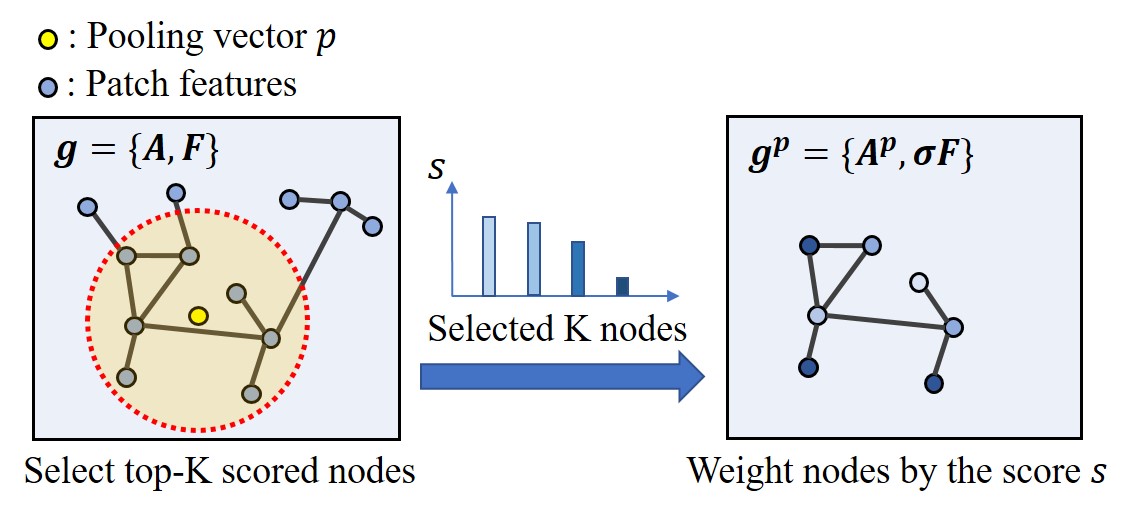}
	\caption{The top-$K$ graph pooling~\cite{graphUnet}. The pooling vector $p$ provides the focused view of the graph for the given task. The final node feature is also weighted by $p$.}
	\label{fig:pooling}
\end{figure}

\begin{figure}[!t]
	\centering
	\includegraphics[width=0.92\linewidth]{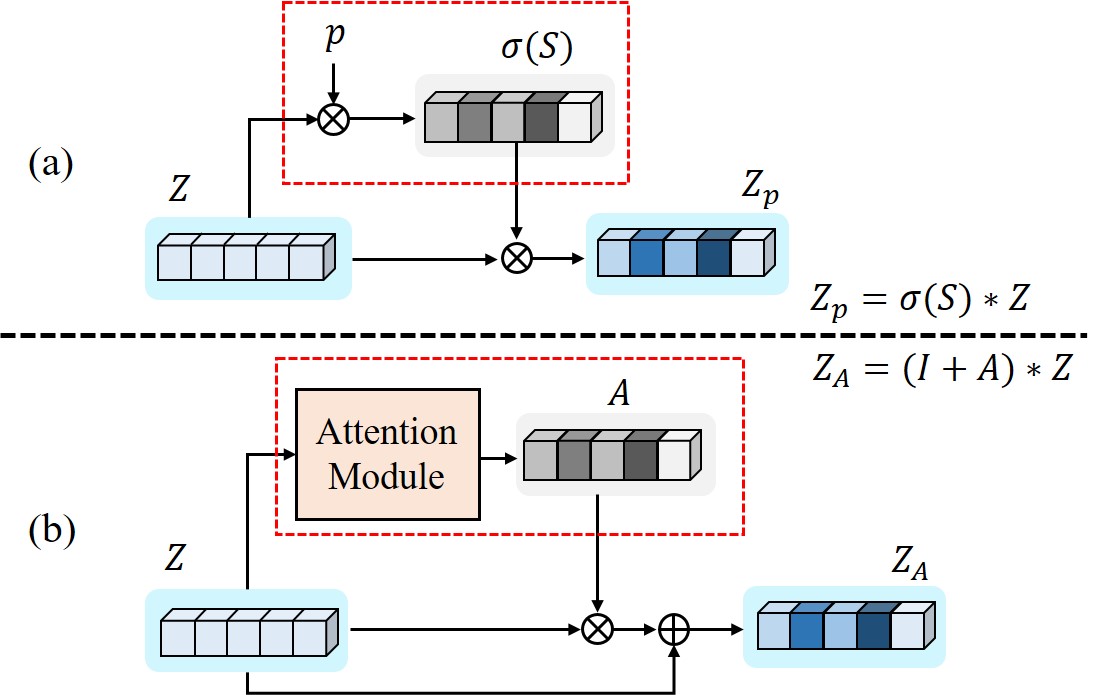}
	\caption{Top-$K$ graph pooling allocates higher weights to the informative nodes, similarly to the attention mechanism. (a) Top-$K$ graph pooling. (b) Attention method.}
	\label{fig:pool-attn}
\end{figure}

Here, it is remarkable how the graph pooling contributes to the improvement. 
As shown in Fig.~\ref{fig:pooling}, the vector $p$ learns to focus on the important nodes, which is determined by the task-relevancy of the nodes. 
It is analogous to the conventional attention methods ~\cite{cbam,bam} shown in Fig.~\ref{fig:pool-attn}. 
Therefore, the graph pooling can be viewed as the node-wise attention, which imposes more regularization for the important nodes to enhance the correspondence for the image translation task.

\paragraph{Overall Loss Function}
Our method is one-sided image translation model without cycle-consistency, inspired by the related works based on the patch-wise contrastive learning ~\cite{HnegSRC, cut, negcut, sesim}. 
Specifically, the overall loss is given as follows:
\begin{align}\label{eq:overall}
	L_{total} &= L_{GAN}(G, D) + \lambda_{g} \sum_{p=0}^P L^p_{GNN}(X, G(X))  \\
	&+ \lambda_{g} \sum_{p=0}^P L^p_{GNN}(Y, G(Y)) \notag
\end{align}
with generator $G$ and discriminator $D$ shown in Fig.~\ref{fig:method}(a). $L_{GAN}$ is LSGAN loss~\cite{lsgan} given as:
\begin{align}
	L_{GAN} = E_{y \sim p_Y} \left[ ||D(y)||_2^2\right] + E_{x \sim p_X} \left[ ||1-D(G(x))||_2^2 \right] 
\end{align}
with the distributions $p_X, p_Y$ for source and target domain.
Additionally, we utilize the identity term $L^p_{GNN}(Y, G(Y))$ to stabilize the training using the target domain images $Y$, as suggested in ~\cite{cut}. $L_{GNN}^{p=0}$ refers the graph loss without the pooling.


\begin{figure*}[!t]
	\centering
	\includegraphics[width=0.97\linewidth]{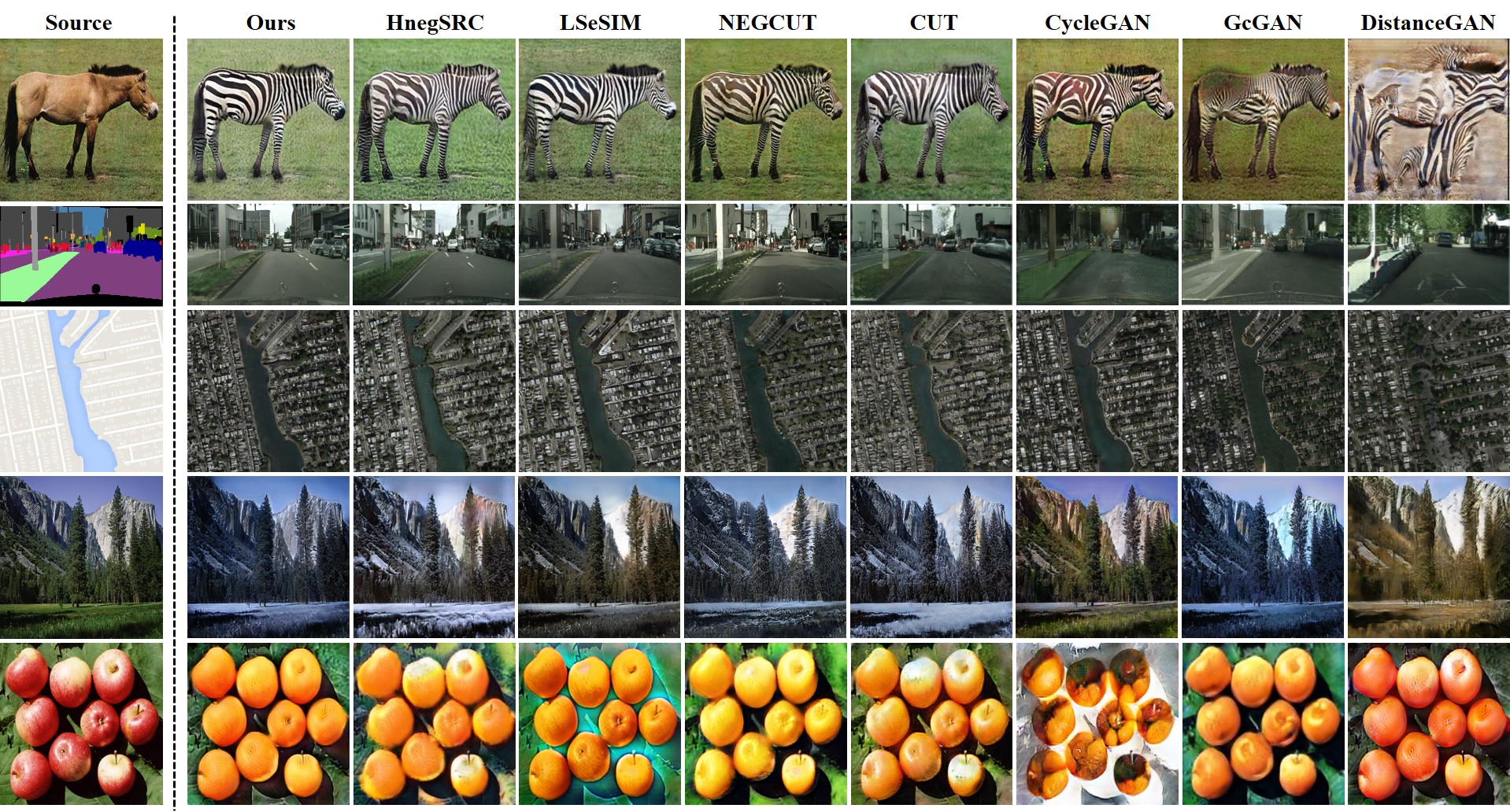}
	\caption{Qualitative comparison with related methods. Our result shows enhanced input-output correspondence, compared to the previous methods.}
	\label{fig:result1}
\end{figure*}

\section{Experimental Results}

\paragraph{Implementation Details}
We first verify our method for unpaired image translation task. 
We verify our method using the five datasets as follows: horse$\rightarrow$zebra, Label$\rightarrow$Cityscape, map$\rightarrow$satellite, summer$\rightarrow$winter, and {apple$\rightarrow$orange}. All images are resized into 256$\times$256 for training and testing. Then, we also present our method for single image translation with high resolution, following the previous work~\cite{cut}. 

For the graph construction, we randomly sampled 256 different patches from the pre-trained VGG16 ~\cite{vgg} network in both of input and output images. We extract the dense feature from the three different layers (relu3-1, relu4-1, relu4-3layer) inside of the network. For the graph operation, we set the number of GNN hops as 2, and pooling number as 1. For the graph pooling, we downsampled nodes by 1/4. In other words, we have 256 nodes in the initial graph, and 64 nodes for the pooled graph. 
More details are provided in the supplementary materials.



\paragraph{Image-to-Image Translation}

We compare our method with the two-sided domain translation models, CycleGAN ~\cite{cyclegan} and MUNIT ~\cite{munit}. Also, we selected the one-sided image translation models, DistanceGAN ~\cite{distancegan} and GcGAN~\cite{gcgan}.
Especially, since our method is based on the patch-wise contrastive learning, we present the comparison with the recent contrastive learning based methods. We compare our method with CUT~\cite{cut} as baseline model, and the improved model of NEGCUT~\cite{negcut}, SeSim~\cite{sesim} and Hneg-SRC ~\cite{HnegSRC}. 

\paragraph{Results}
\begin{figure}[!t]
	\centering
	\includegraphics[width=0.85\columnwidth]{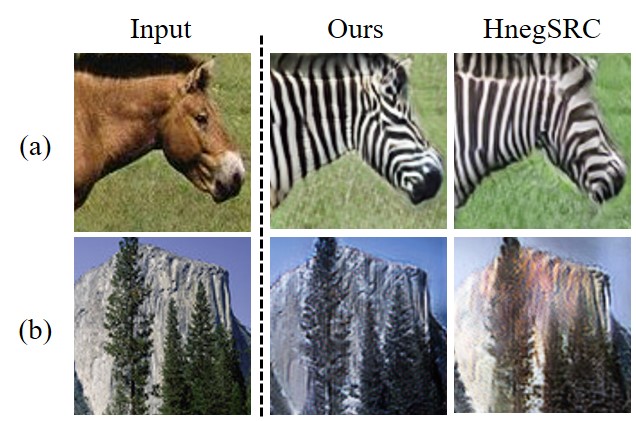}
	\caption{Closer views of the output images. Our method enhances the spatial-specific information given in the input.}
	\label{fig:spatial_specific}
\end{figure}


The results in Fig.~\ref{fig:result1} verifies that the proposed method generates the images with better visual quality than the other methods, by enhancing the correspondence between the input and the output images. Compared to the other methods, our methods preserves the structural information of the input images, by using the patch-wise connection of the input as the prior knowledge. 

Moreover, we further compare our method with the HnegSRC which also utilizes the patch-wise semantic relation of the input. As shown in Fig.~\ref{fig:spatial_specific}, our method enhances the spatial-specific information considering the patch-wise semantic neighborhood by the graph operation, compared with the HnegSRC which only imposed the consistency regularization for the patch-wise similarity. Specifically, our method in Fig.~\ref{fig:spatial_specific}(a) outputs more realistic zebra by showing the spatial-specific patterns(e.g. dark colored mouth), which is not in the compared result. Also, our result in Fig.~\ref{fig:spatial_specific}(b) shows the tree branches with the coherent shapes to the input, which are distorted in the compared method.  

The results in Table~\ref{table:main} also supports the outperformance of the proposed method. 
Specifically, in horse$\rightarrow$zebra and Label$\rightarrow$Cityscape datasets, we similar FID scores with the HnegSRC, but higher scores by KID. 
For summer$\rightarrow$winter and apple$\rightarrow$orange datasets, our model outperformed the others by large margins, which demonstrates the effectiveness of the proposed model.

\begin{table*}[!t]
\begin{center}
\resizebox{0.93\textwidth}{!}{
	
	\begin{tabular}{@{\extracolsep{2pt}}ccccccccccc@{}}
		\hline
		\multirow{2}{*}{\textbf{Method}}  & \multicolumn{2}{c}{\textbf{Horse$\rightarrow$Zebra}} & \multicolumn{2}{c}{\textbf{Label$\rightarrow$Cityscape}}&\multicolumn{2}{c}{\textbf{Map$\rightarrow$Satellite}}&\multicolumn{2}{c}{\textbf{Summer$\rightarrow$Winter}}&\multicolumn{2}{c}{\textbf{Apple$\rightarrow$Orange}}\\
		
		\cline{2-3} 
		\cline{4-5}
		\cline{6-7} 
		\cline{8-9}
		\cline{10-11}
		
		& FID$\downarrow$&KID$\downarrow$& FID$\downarrow$&KID$\downarrow$& FID$\downarrow$&KID$\downarrow$& FID$\downarrow$&KID$\downarrow$ &FID$\downarrow$&KID$\downarrow$\\
		
		\hline
		CycleGAN &77.2&	1.957 &76.3 & 3.532 &54.6&3.430	&84.9&	1.022&	174.6&	10.051\\
		MUNIT &133.8&3.790  & 91.4&6.401 &181.7	&12.03&115.4&	4.901&	207.0&	12.853\\
		DistanceGAN & 72.0& 1.856 & 81.8& 4.410&98.1&	5.789&97.2&	2.843&	181.9&	11.362\\
		GCGAN &86.7&	2.051&105.2	&6.824&79.4&	5.153&97.5	&2.755&	178.4&	10.828\\
		\hdashline
		CUT & 45.5&	0.541	& 56.4&1.611 &56.1	& 3.301&84.3	&1.207&	171.5&	9.642\\
		NEGCUT& 39.6&	0.477& 48.5& 1.432&	51.0&	2.338&82.7&1.352&	154.1&	7.876\\
		LSeSIM & 38.0 &	0.422&	49.7&2.867&52.4	&3.205&83.9&	1.230&	168.6&	10.386\\
		HnegSRC & \textbf{34.4}&	0.438&	\textbf{46.4} & 0.662 &49.2& 	2.531&81.8	&1.181&	158.3&	8.434\\
		\hdashline
		Ours & 34.5&	\textbf{0.271}&	46.8& \textbf{0.605}& \textbf{45.9}&	\textbf{2.112}&\textbf{75.8}	& \textbf{0.845}&	\textbf{139.1}&	\textbf{7.134}\\
		\hline
		
	\end{tabular}
}
\end{center}
\caption{Quantitative results. Our model outperforms the baselines in both of FID and KID$\times$100 metrics. }

\label{table:main}
\end{table*}

\begin{figure*}[!t]
	\centering
	\includegraphics[width=0.88\linewidth]{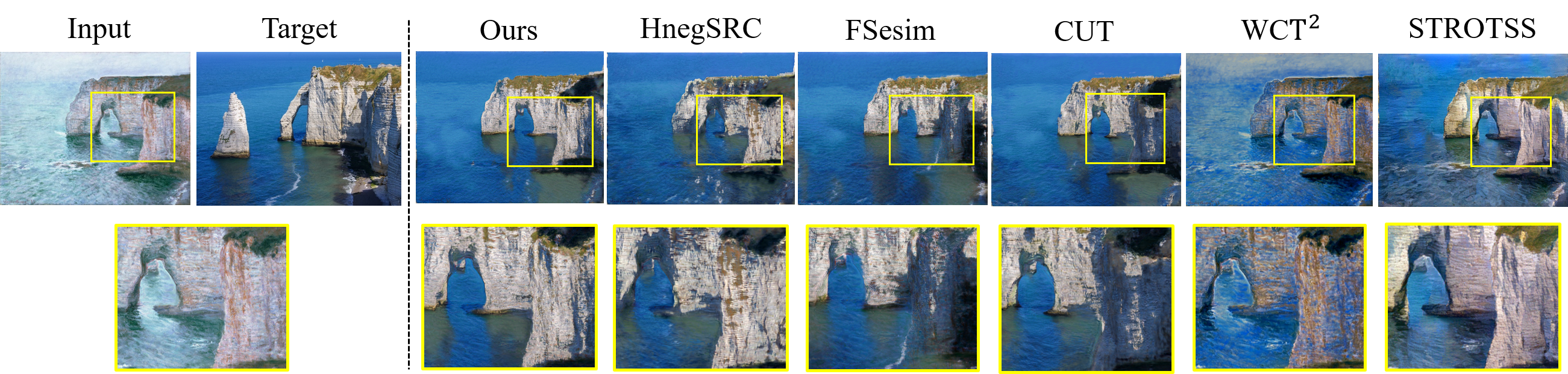}
	\caption{Qualitative comparison on single image translation.}
	\label{fig:single}
\end{figure*}

\begin{figure*}[t]
	\centering
	\includegraphics[width=0.83\linewidth]{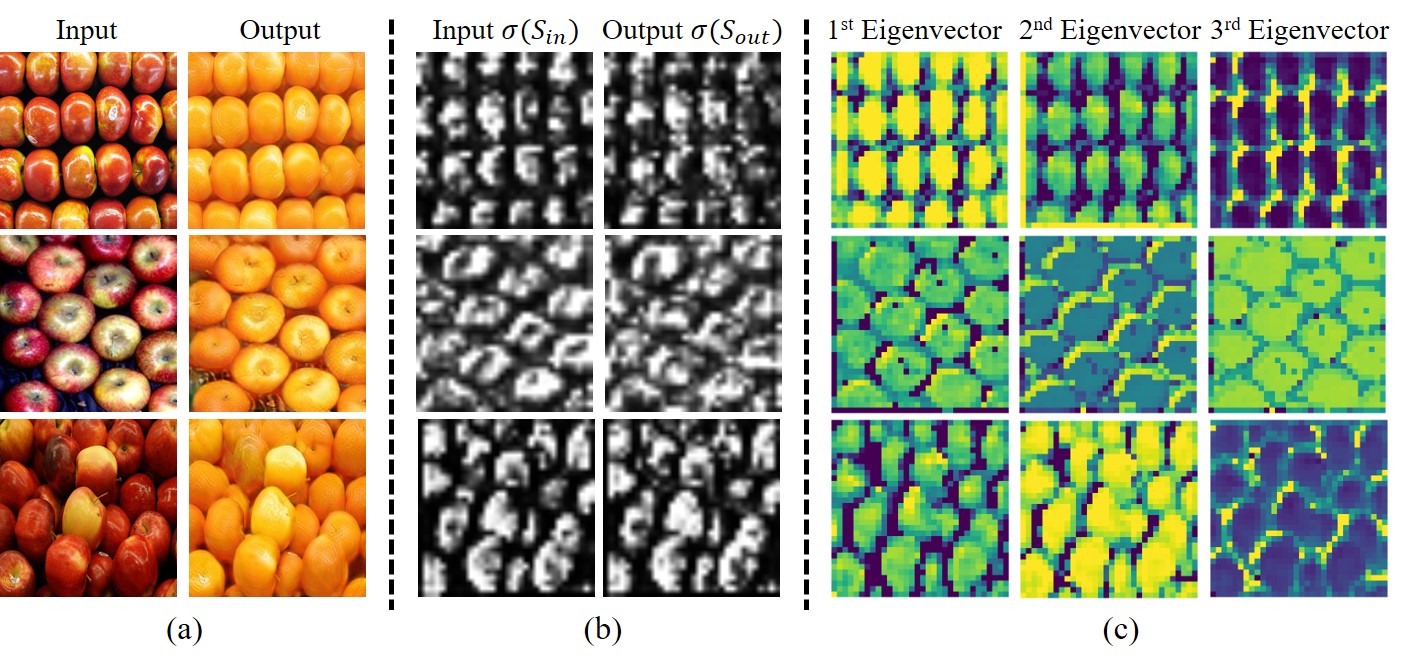}
	\captionof{figure}{Analysis of the proposed method: (a) Input and the output images. (b) Visualization of $\sigma(S_{in}), \sigma(S_{out})$. The vector $p$ allocates higher weights for the object parts which are task-relevant. Similar appearance refers the correspondence between input and output. (c) Eigenvectors of the Laplacian matrix of $A$, which are coherent to the semantics of the image.} 
	\label{fig:analy}
\end{figure*}

\begin{figure}[!t]
	\centering
	\includegraphics[width=0.98\linewidth]{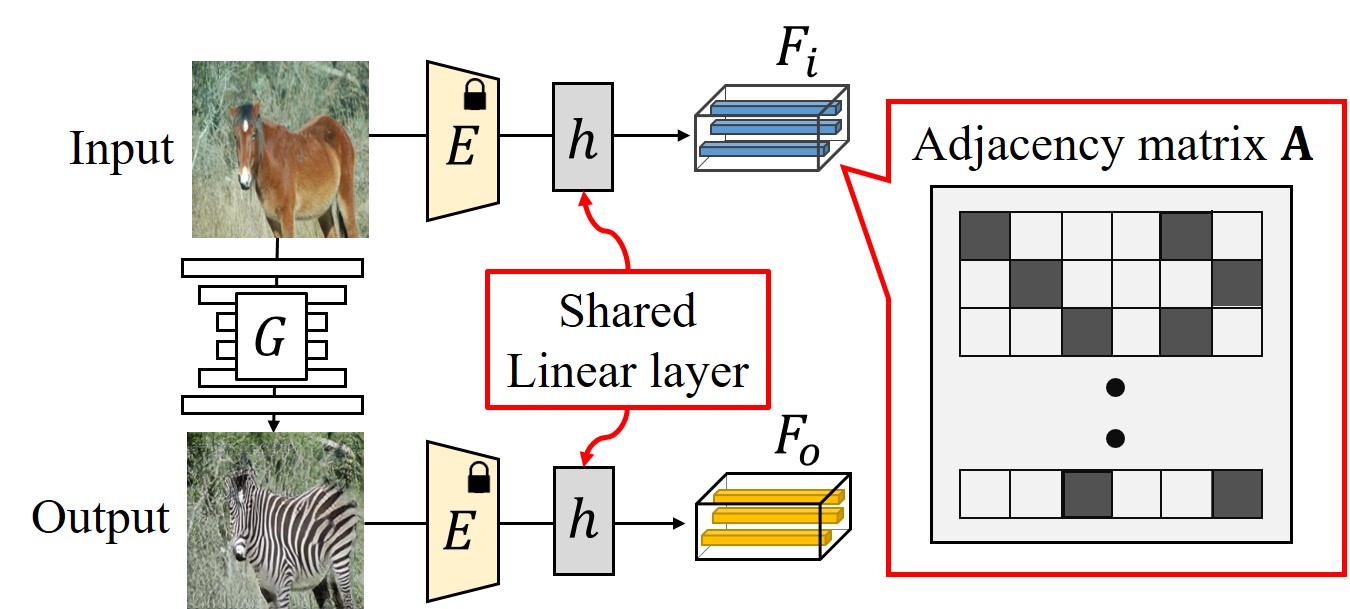}
	\caption{The adjacency matrix $A$ is constructed from $F_i$ which is the output of learnable $h$. Here, $h$ is updated by the gradient from the $F_o$ similar to CUT~\cite{cut}. } 
	\label{fig:adj_FC}
\end{figure}

\paragraph{Single-Image Translation}
Following the previous work~\cite{cut}, we verify our method for the single image translation. The input is a Claude Monet's painting, and the target domain is the natural landscape images. 
Detailed experimental settings are provided in the supplementary material. 
For comparison, we choose previous methods, STROTSS ~\cite{strotss} and WCT2 ~\cite{wct2}. Also, we selected the patch-based contrastive learning methods, which are CUT ~\cite{cut}, FSeSim ~\cite{sesim} and HnegSRC ~\cite{HnegSRC}.  

Fig.~\ref{fig:single} shows the qualitative comparison of the single-image translation. To show the detailed visual comparison, we enlarge a specific region which is annotated as the yellow box. In WCT2 and STROTSS, the outputs are not fully changed, which contains the artistic textures of the input. The contrastive learning based methods output the improved results, however, show some deformation in shapes. 
Compared to the previous methods, our method generates realistic image with the enhanced correspondence to the input. 

\section{Discussion}

The proposed method consists of two parts. First, we construct the graph by the pretrained encoder. Second, we utilize top-$K$ pooling by the pooling vector $p$ to focus on task-relevant nodes which provides the localized graph. 
To investigate the effectiveness of each part, we first investigate what the vector $p$ learns for the graph pooling procedure. 
Second, we investigate the adjacency matrix $A$ in Fig.~\ref{fig:adj_FC}
to verify the patch-wise connection used to construct the graph.

%

\paragraph{Semantic Meaning of the Pooling Vector $p$ }
Recall that the vector $p$ allocates higher weights to focus on the important nodes of the graph, which is analogous to the attention mechanism. 
Here, we provide empirical results which indicates how the vector $p$ allocates weights for nodes $Z, V$.

Specifically, we visualize $\sigma(S_{in}), \sigma(S_{out})$, given by:
\begin{align}
	S_{in} &= p^\top Z \\
	S_{out} &= p^\top V
\end{align}
where the $\sigma$ is sigmoid function.
From the result in Fig.~\ref{fig:analy}(b), we can derive two main points. 
First, the vector $p$ focuses mainly on the object patches which are semantically close and task-relevant. 
Considering that the top $K$ nodes are selected in graph pooling, the result verifies that the vector $p$ provides focused view of graph by selecting informative nodes.
Second, we can observe that the focused parts in $\sigma(S_{in}), \sigma(S_{out})$ are similar. Therefore, the node features $Z, V$ are semantically coherent, indicating the correspondence between the input and the output images.

\paragraph{Adjacency Matrix $A$}
As shown in Fig.~\ref{fig:adj_FC}, 
we construct the graph by the learnable adjacency matrix $A$ obtained from the feature $F_i$, which is the output of the learnable layer $h$ as shown in Fig.~\ref{fig:pooling}. We visualize the eigenvectors of the graph Laplacian matrix to verify the learned patch-wise connection in the graph, as suggested in ~\cite{deepSpectral}. 

Fig.~\ref{fig:analy}(c) shows that the eigenvectors are semantically coherent with the input image, which clearly demonstrates that the adjacency matrix captures the appropriate implicit semantic connection of the given image.

\paragraph{Ablation Study for Graphs}
We provide the ablation study on the graph in Table~\ref{table:ablation}, regarding on the number of hops~($n$), value for similarity threshold~($t$), number of graph pooling layer and downsampling ratio of the pooling. 
First, for the ablation study on GNN, we observe that results are degraded for both lower ($n=1$) and larger number of hops ($n=3$). Also, our setting shows the best result compared to the other threshold values~($t=0.0, 0.4, 0.6$). 
Especially for the increased $t$ (i.e. sparse connectivity), the model shows much degradation. 
This suggests that sufficiently dense graph captures meaningful topology.

Second, for the ablation study on pooling layers, the performance is degraded without the pooling layer (\# of pool=0), as the network do not leverage the information from the focused view. For more pooling layers, models also show degraded performance, as the pooled graphs have fewer nodes which leads to fewer negative pairs for the contrastive learning. Additionally, we provide the results with varying downsampling rate. For 1/8 downsampling, the pooled graph have fewer nodes, which leads to similar problem with the excessive pooling layers. 
This again confirms that a sufficiently dense graph after the pooling can capture the semantically meaningful hiearchy. 
We provide additional ablation study for the graph construction in the supplementary material.

\begin{table}[!t]
	\begin{center}
		\resizebox{0.47\textwidth}{!}{
			\begin{tabular}{@{\extracolsep{1.5pt}}ccccccc@{}}
				\hline
				
				\multicolumn{5}{c}{\textbf{Settings}} & \multicolumn{2}{c}{\textbf{H$\rightarrow$Z}}  \\ 
				\cline{1-5} 
				\cline{6-7} 
				& Hops & {Thresh.} &\# of& Down&\multirow{2}{*}{FID$\downarrow$}& \multirow{2}{*}{KID$\downarrow$} \\ 
				&($n$)& ($t$) & Pool &   sample & &  \\
				\cline{1-5} 
				\cline{6-7} 
				\multirow{5}{*}{\parbox{0.72cm}{GNN ($n, t$)}} &1&0.1&1&1/4 &37.9 &0.438\\ 
				&3&0.1&1&1/4& 39.9 & 0.374 \\ 
				\cdashline{2-7}	
				&2&0.0&1&1/4 & 34.5& 0.551\\ 
				&2&0.4&1&1/4 & 36.8& 0.293\\ 
				&2&0.6&1&1/4 & 38.3&0.332\\ 
				\hline
				\multirow{3}{*}{\parbox{0.6cm}{Graph Pool}} &2&0.1&0&- & 37.6&0.432\\ 
				&2&0.1&2&1/4 & 35.0 &0.625\\ 
				&2&0.1&1&1/8& 37.7&0.340\\
				\hline
				\textbf{Ours}&\textbf{2}&\textbf{0.1}&\textbf{1}&\textbf{1/4}& \textbf{34.5}&\textbf{0.271}  \\
				\hline				
			\end{tabular}
		}
	\end{center}
	\caption{Quantitative results of ablation studies. Our setting shows the best performance in both of FID and KID$\times$100.}
	\label{table:ablation}
\end{table}

\section{Conclusion}
In conclusion, we proposed a novel patch-wise graph representation matching method for image translation task. For structural consistency between input and output images, we proposed to match the constructed graphs between input and outputs. In this part, we used the same adjacency matrix for input and output images for graph consistency. To further leverage the topological information in an hierarchical manner, we applied graph pooling on initial graphs. Our experimental results showed state-of-the-art performance, which again confirms that graph-based patch representation have obvious advantage over baseline methods.


\section{Ethical Statement}
Regarding on the social impact, the realistic fake images generated by the proposed method may produce a social disinformation, as most of image generation methods shares. Also, the model has potential risk of violating copyright as the model learns the mapping function from input to target distribution.

\section{Acknowledgements}
This research was supported by National Research foundation of Korea(NRF) (**RS-2023-00262527**). This research was also supported by Field-oriented Technology Development Project for Customs Administration through National Research Foundation of Korea(NRF) funded by the Ministry of Science \& ICT and Korea Customs Service(**NRF-2021M3I1A1097938**).

\bibliography{aaai24}

\end{document}